\documentclass[pre,twocolumn,twoside,showpacs,superscriptaddress,floatfix]{revtex4-1}
\usepackage{graphics,amssymb,amsmath,epsf,color,hyperref}
\epsfclipon
\usepackage{epsfig,bm,epsf,graphics}
\usepackage{graphicx}
\usepackage{dcolumn}
\usepackage{bm}
\usepackage{braket}
\usepackage{color}
\usepackage{soul}
\usepackage{xcolor}

\begin{document}

\title{Information Flows of Diverse Autoencoders}

\author{Sungyeop Lee}
\email[Corresponding author: ]{dtd2001@snu.ac.kr}
\affiliation{Department of Physics and Astronomy, Seoul National University, Seoul 08826, Korea}

\author{Junghyo Jo}
\email[Corresponding author: ]{jojunghyo@snu.ac.kr}
\affiliation{Department of Physics Education and Center for Theoretical Physics and Artificial Intelligence Institute, Seoul National University, Seoul 08826, Korea}
\affiliation{School of Computational Sciences, Korea Institute for Advanced Study, Seoul 02455, Korea}

\date{\today}

\begin{abstract}
Deep learning methods have had outstanding performances in various fields. 
A fundamental query is why they are so effective. Information theory provides a potential answer by interpreting the learning process as the information transmission and compression of data. The~information flows can be visualized on the information plane of the mutual information among the input, hidden, and output layers. In this study, we examine how the information flows are shaped by the network parameters, such as depth, sparsity, weight constraints, and hidden representations.
Here, we adopt autoencoders as models of deep learning, because (i) they have clear guidelines for their information flows, and (ii) they have various species, such as vanilla, sparse, tied, variational, and~label autoencoders. 
We measured their information flows using   R{\'e}nyi's matrix-based $\alpha$-order entropy functional. As learning progresses, they show a typical fitting phase where the amounts of input-to-hidden and hidden-to-output mutual information both increase. In the last stage of learning, however, some~autoencoders show a simplifying phase, previously called the “compression phase”, where input-to-hidden mutual information diminishes. In particular, the sparsity regularization of hidden activities amplifies the simplifying phase. However, tied, variational, and label autoencoders do not have a simplifying phase. Nevertheless, all autoencoders have similar reconstruction errors for training and test data. Thus, the simplifying phase does not seem to be necessary for the generalization of learning.
\end{abstract}

\maketitle


\section{Introduction}

Since the development of information theory as a theory of communication by Shannon~\cite{Shannon, Cover}, it has played a crucial role in various domains of engineering and science, including physics~\cite{Jaynes}, biology~\cite{Yockey}, and machine learning~\cite{Mackay}.
The information bottleneck (IB) theory interprets the learning process of neural networks as the transmission and compression of information~\cite{Tishby}.
Neural networks encode input $X$ into internal representation $Z$; then, they decode $Z$ to predict the desired output $Y$.
The IB theory is a rate-distortion theory that compresses irrelevant information in $X$ for predicting $Y$ to the maximum extent.
The objective function of this theory can be mathematically described as minimizing the mutual information $I(X;Z)$ between $X$ and $Z$, given a required transmission $I_{req}$ of the mutual information $I(Z;Y)$ between $Z$ and $Y$:
\begin{eqnarray}
\label{eq:IB}
  \min_{p(Z|X)} I(X;Z)- \beta \Big[ I(Z;Y) - I_{req} \Big],
\end{eqnarray}
where $\beta$ is a trade-off coefficient that balances the information compression and transmission. {The numerical method for solving this optimization problem, so called Blahut--Arimoto algorithm~\cite{Arimoto, Blahut}, has been extensively studied using various problems~\cite{BA_app1, BA_app2}. Theoretical} aspects of IB theory and its applications are well summarized in~\cite{IB_review}.

Here, it is important to note that machine learning models, including ours in this study, do not take Equation~(\ref{eq:IB}) as the loss function, although new deep variational IB models directly adopt it as the loss function~\cite{Alemi2017}. The IB theory provides a nice interpretation of learning process, but it does not work for the optimization of neural networks in general.
The mutual information provides a potent tool for visualizing the learning processes by displaying a trajectory on the two-dimensional plane of $I(X;Z)$ and $I(Z;Y)$, called~the {information plane} (IP).
Through IP analyses, Shwartz-Ziv and Tishby found that the training dynamics of neural networks demonstrate a transition between two distinct phases: fitting and compression~\cite{IB, IP}. 
Supervised learning experiences a short fitting phase in which the training error is significantly reduced. This first phase is characterized by  increases in $I(X;Z)$ and $I(Z;Y)$.
Then, in the learning process, a large amount of time is spent on finding the efficient internal representation $Z$ of input $X$ when the fitting phase secures a small training error.
During this second phase of compression, $I(X;Z)$ decreases while $I(Z;Y)$ remains constant.
To avoid unnecessary confusion with the usual data compression or dimensionality reduction, henceforth, we denote the second phase as “simplifying phase” instead of the original name of “compression phase”.

The simplifying phase has been argued to be associated with the generalization ability of machine learning by compressing irrelevant information of training data to prevent overfitting~\cite{IP}. 
The non-trivial simplifying phase and its association with generalization have been further observed in other studies using different network models with different data; however, the universality of the simplifying phase remains debatable~\cite{generalIP, IP&initialization, IP&CNN}.
The~debates can be partly attributed to the sensitivity toward the architecture of neural networks, activation functions, and estimation schemes of information measures.

In this study, we investigate how the information flows are shaped by the network designs, such as depth, sparsity, weight constraints, and hidden representations, by using autoencoders (AEs) as specific models of machine learning. 
AEs are neural networks that encode input $X$ into internal representation $Z$ and reproduce $X$ by decoding $Z$. 
This~representation learning can be interpreted as self supervised learning where a label is input as itself, such as $Y = X$.
To examine the IP analyses of representation learning, we considered AEs because (i) they have a concrete guide ($Y = X$) for checking the validity of $I(X;Z)$ and $I(Z;Y)$ on the IP, (ii) they have various species to fully explore trajectories on the IP, and~(iii) they are closely related to unsupervised learning.

The remainder of this paper is organized as follows. We introduce various types of AEs in Section~\ref{autoencoders} and explain our matrix-based kernel method for estimating mutual information in Section~\ref{MI}. Then, we examine the IP trajectories of information transmission and compression of the AEs in Section~\ref{results}. Finally, we summarize and discuss our findings in Section~\ref{discussions}.
 
\section{Representation Learning in Autoencoders}
\label{autoencoders}
\subsection{Information Plane of Autoencoders}
AEs are neural networks specialized for dimensional reduction and representation learning in an unsupervised manner. A deep AE consists of a symmetric structure with encoders and decoders as follows:
\begin{eqnarray}
    X-E_1-\cdots-E_L-Z-D_1-\cdots-D_L-X'.
\end{eqnarray}
where $E_i$ and $D_i$ denote the $i$-th encoder and decoder layer, respectively, and $Z$ is the bottleneck representation with the smallest dimension. The deep AE trains an identity function to reproduce input $X$ from output $X'$.
During the training process, the AE extracts relevant features for reproducing $X$ while compressing the high-dimensional input $X$ into an internal representation $Z$ on a low-dimensional bottleneck layer. 
The encoder and decoder layers form Markov chains that should satisfy the {data processing inequality} (DPI), analogously to supervised learning~\cite{IP&AE1}:
\begin{align}
    \textrm{Forward DPI:} I(X;E_1)\geq \cdots \geq I(X;E_L)\geq I(X;Z),\\
    \textrm{Backward DPI:} I(Z;X')\leq I(D_1;X')\leq \cdots \leq I(D_L;X').
\end{align}

The forward DPI represents information compression as input $X$ is processed into the bottleneck layer, whereas the backward DPI represents information expansion as the compressed representation $Z$ is transformed into output $X'$. 
It is noteworthy that the usual AEs have physical dimensions, narrowing toward the bottleneck and expanding away from the bottleneck, which are consistent with the DPI.

The desired output of this AE is identical to the input ($X' = X$).
This identity constrains the input and output mutual information to be located on a straight line $I(X;T)=I(T;X)$ for arbitrary internal representations, $T \in \{E_1, \cdots, E_L, Z, D_1, \cdots, D_L \}$.
Here, if the desired output $X$ in $I(T;X)$ is replaced with the predicted output $X'$ of the AE, the learning dynamics of the AE on the IP can be analyzed~\cite{IP&AE1}.
Then, the two sets of mutual information for representing information compression and transmission correspond to
\begin{eqnarray}
    I(X;T)&=H(T)-H(T|X)\\
    I(T;X')&=H(T)-H(T|X'),
\end{eqnarray}
where $H(T)$ represents the Shannon entropy of $T$, and $H(T|X)$ and $H(T|X')$ are the conditional entropies of $T$ given $X$ and $X'$, respectively.
The forward process of the AE realizes the deterministic mapping of $T$ = $T(X)$ and $X'$ = $X'(T)$. Then, one-to-one correspondence of $X \rightarrow T$ implies no uncertainty for $H(T(X)|X) = 0$, whereas the possibly many-to-one correspondence of $T \rightarrow X'$ implies some uncertainty for $H(T|X'(T)) \neq 0$.
Therefore, the~inequality of $I(X;T) \geq I(T;X')$ is evident because $H(T) \geq H(T) - H(T|X')$, where~the conditional entropy $H(T|X')$ is non-negative.
Based on this inequality, the learning trajectory of $I(X;T)$ and $I(T;X')$ on the two-dimensional IP ($x, y$) can be expected to stay below the diagonal line $y = x$.
Once the learning process of the AE is complete with $X'=X$, the~two sets of mutual information become equal to $I(X;T) = I(T;X'=X)$, and~the learning trajectory ends up on the diagonal line.

\subsection{Various Types of Autoencoders}
To investigate information flows of machine learning, we adopted AEs because their theoretical bounds of IP trajectories could guide our IP analysis. 
IP analysis has been used to visualize the information process in AEs~\cite{IP&AE1, IP&AE2}.
Their IP trajectories satisfied the theoretical boundary of $I(X;T) \geq I(T;X')$. Previous studies examined IP trajectories according to the size of the given bottleneck layers, but they did not investigate the associations between the simplifying phases and generalizations of AEs.
Another important advantage of adopting AEs is their diverse variants that enable us to fully explore IP trajectories depending on the network designs, such as depth, sparsity, weight constraints, and hidden representations. In particular, because a certain AE model is directly linked to unsupervised learning, the~model can be used to understand the information process of unsupervised learning.
Now, we briefly introduce diverse species of AEs used in our experiments.

The simplest structure of AE, called shallow AE, consists of a single bottleneck layer between the input and output layers. In the shallow AE ($X-Z-X'$), the forward propagation of input $X$ is defined as
\begin{eqnarray}
    Z &= f_E(W_EX+b_E)\\
    X' &= f_D(W_DZ+b_D),
\end{eqnarray}
where $W$ and $b$ represent the weights and biases, respectively, and $f(s)$ is a corresponding activation function. Here, the subscripts $E$ and $D$ denote the encoder and decoder, respectively.
The shallow AE is trained to minimize the reconstruction errors usually measured by the mean squared error (MSE) between output $X'$ and desired output $X$. It has been analytically proven that a shallow AE with linear activation ($f(s)=s$) spans the same subspace as that spanned by principal component analysis (PCA)~\cite{AE_PCA1, AE_PCA2}. Deep AEs stack hidden layers in the encoder and decoder symmetrically; moreover, it is well known that deep AEs yield better compression than shallow AEs. 

Right up till recently, a myriad of variants and techniques have been proposed to improve the performance of AEs via richer representations, such as sparse AE (SAE)~\cite{SAE}, tied AE (TAE)~\cite{DAE}, variational AE (VAE)~\cite{VAE}, and label AE (LAE)~\cite{SemanticAE, SupervisedAE}. 
\begin{itemize}
\item
SAE was proposed to avoid overfitting by imposing sparsity in the latent space. The~sparsity penalty is considered a regularization term of the Kullback--Leibler (KL) divergence between the activity of bottleneck layer $Z$ and sparsity parameter $\rho$, a~small value close to zero. 

\item
TAE shares the weights for the encoder and decoder part ($W_E=W_D^T$), where superscript $T$ depicts the transpose of a matrix.
This model is widely used to reduce the number of model parameters while maintaining the training performance. Owing to its symmetrical structure, it can be interpreted as a deterministic version of restricted Boltzmann machines (RBMs), a representative generative model for unsupervised learning; consequently, the duality between TAE and RBM has been identified~\cite{AE_RBM}. Compared to the vanilla AE, SAE and TAE have regularizations for the degrees of freedom for nodes and weights, respectively. 
 Later, we visually validate how these constraints lead to a difference in the information flow of IP trajectories.

\item
The ultimate goal of AEs is to obtain richer expressions in the latent space. 
Therefore, an AE is not a mere replica model, but a generative model that designs a tangible latent representation to faithfully reproduce the input data as output.
VAE is one of the most representative generative models with a similar network structure to AE; however, its mathematical formulation is fundamentally different. 
The detailed derivation of the learning algorithm of VAE is beyond the scope of this study, and thus it will be omitted~\cite{VAE}.
In brief, the encoder network of VAE realizes an approximate posterior distribution $q_\phi(Z|X)$ for variational inference, whereas the decoder network realizes a distribution $p_\theta(X|Z)$ for generation. The loss of VAE, known as the evidence lower bound (ELBO), is decomposed into a reconstruction error given by the binary cross entropy (BCE) between the desired output $X$ and predicted output $X'$, and~the regularization of KL divergence between the approximate posterior distribution $q_\phi(Z|X)$ and prior distribution $p(Z)$.  
As tangible Gaussian distributions are usually adopted as the approximate posterior and prior distributions of $q_\phi(Z|X)$ and $p(Z)$, respectively, VAE~has a special manifold of the latent variable $Z$.
 
\item
AEs do not use data labels.
Instead, inputs work as self labels for supervised learning.
Here, to design the latent space using label guides, we consider another AE, called~label AE (LAE).
LAE forces the input data to be mapped into the latent space with the corresponding label classifications.
Then, the label-based internal representation is decoded to reproduce input data.
 Although the concept of regularization using labels has been proposed~\cite{SemanticAE, SupervisedAE}, LAE has not been considered as a generative model.
\mbox{Unlike}~vanilla AEs that use a sigmoid activation function, LAE uses a softmax activation function, $f(Z_i)=\exp (Z_i)/{\sum_j\exp(Z_j)}$, to impose the regularization of the internal representation $Z$ to follow the true label $Y$ as the cross entropy (CE) between $Y$ and $Z$.
Once LAE is trained, it can generate learned data or images using its decoder, starting from one-hot vector $Z$ of labels with the addition of noise.
Additional details of LAE are provided in Appendix \ref{appendixB}. 
Later, we compare the IP trajectories of VAE and LAE with those of vanilla AE in a deep structure to examine how the information flow varies depending on the latent space of generative models. 
\end{itemize}

Table \ref{AE type} summarizes the loss function, constraints, and activation function of the bottleneck layer for each aforementioned AE model.

\begin{table}[h] 
\caption{Various species of autoencoder. Vanilla AE uses the mean squared error (MSE) loss. We used a sigmoid function as an activation function for the bottleneck layer, which helps in unifying the scales of different layers. Regularization of SAE is the KL-divergence between the hidden activity and sparsity parameter $\rho$. The only difference in TAE is that it shares the weight of encoder and decoder. The loss function of VAE, known as the evidence lower bound (ELBO), consists of the reconstruction error, binary cross entropy (BCE), and KL-divergence between the approximate posterior $q_{\phi}(Z|X)$ and prior $p(Z)$; moreover, the stochastic node activities of the bottleneck layer are sampled from Gaussian distributions. In LAE, the classification error, the cross entropy (CE) between the softmax hidden activity $Z$ and true label $Y$, is used as a regularization term.}
    \begin{tabular}{cccc}
    \hline
         {Model} & {Main Loss} & {Constraint} & {Bottleneck Activation}\\
         \hline
         AE & MSE$(X,X')$ & None & sigmoid \\
         SAE & MSE$(X,X')$ & KL$(\rho||Z)$ & sigmoid \\
         TAE & MSE$(X,X')$ & $W_E = W_D^T$ & sigmoid \\
         VAE & BCE$(X,X')$ & KL$(q_\phi(Z|X)||p(Z))$ & Gaussian sampling \\
         LAE & MSE$(X,X')$ & CE$(Y,Z)$ & softmax \\
         \hline
    \end{tabular}
\label{AE type}
\end{table}

\section{Estimation of Mutual Information}
\label{MI}
After preparing various species of AE models to explore diverse learning paths on the IP, we need to estimate the mutual information for IP analyses:
\begin{eqnarray}
    I(X;Z) = \sum_{x,z} p(x,z) \log \frac{p(x,z)}{p(x) p(z)}. \label{def_MI}
\end{eqnarray}

In reality, we have samples of data, $\{x(t), z(t) \}_{t=1}^N$, instead of their probabilities, $p(x),$ $p(z)$, and $p(x,z)$.
Using $N$ samples of data, we may estimate the probabilities. Here, if~$X$ and $Z$ are continuous variables, it is inevitable to first discretize them.
Then, we can count the discretized samples for each bin and estimate the probabilities.
The~estimation of mutual information based on this binning method has some limitations.
First, its accuracy depends on the resolution of discretization. Second, large samples are required to properly estimate the probability distributions. Suppose that $X$ is an $n$-dimensional vector. Despite considering the most naive discretization with binarized activities, the total number of configurations for the binarized $X$ is already $2^n$. Thus it becomes impracticable for $N$ finite samples to cover the full range of configurations, e.g., $2^{20} \approx 10^6$ configurations for $n=20$.    
Therefore, other schemes, such as kernel density estimation~\cite{KDE}, k-nearest neighbors, and~matrix-based kernel estimators~\cite{kernel, Yu2019}, exist to estimate the entropy and mutual information. The description of each scheme and the corresponding IP results were presented in a pedagogical review~\cite{Review}. Among these various methods, we adopted a matrix-based kernel estimator, which is mathematically well defined and computationally efficient for large networks. It estimates the R\'enyi's $\alpha$-order entropy using the eigenspectrum of covariance matrix of $X$ as follows: 
\begin{align}
    S_\alpha(A)=\frac{1}{1-\alpha}\log_2\left[\text{tr}(A^\alpha)\right]=\frac{1}{1-\alpha}\log_2\left[\sum_{i=1}^N\lambda_i(A)^\alpha\right], \label{Renyi entropy}
\end{align}
where $A$ is an $N\times N$ normalized Gram matrix of random variable $X$ with size $N$ and $\lambda_i(A)$ is the $i$-th eigenvalue of $A$.
Note that $\text{tr}$ denotes the trace of a matrix.
In the limit of $\alpha \rightarrow 1$, Equation~(\ref{Renyi entropy}) is reduced to an entropy-like measure that resembles the Shannon entropy of $H(X)$. If we assume that $B$ is a normalized Gram matrix from another random variable $Z$, the joint entropy between $X$ and $Z$ is defined as
\begin{eqnarray}
    S_\alpha(A,B)=S_\alpha\left(\frac{A\circ B}{\text{tr}(A\circ B)}\right), \label{joint entropy}
\end{eqnarray}
where $A\circ B$ denotes the Hadamard product, i.e., the element-wise product of two matrices. 
From Equations~(\ref{Renyi entropy}) and (\ref{joint entropy}), the mutual information can be defined as
\begin{eqnarray}
    I_\alpha(X;Z)=S_\alpha(A)+S_\alpha(B)-S_\alpha(A,B), \label{mutual information}
\end{eqnarray}
which is analogous to the standard mutual information in the new space called reproducing kernel Hilbert space (RKHS).
Although $I_\alpha(X;Z)$ is mathematically different from  $I(X;Z)$ in Equation~(\ref{def_MI}), this quantity satisfies the mathematical requirements as R\'enyi's entropy~\cite{kernel}. Furthermore, it has a great computational merit in that its computation is not affected much by the dimensiosn $n$ of $X$, unlike the standard binning method for estimating the mutual information.
In a simple setup where an exact computation of $I(X;Z)$ is possible, we confirmed that the matrix-based $I_\alpha(X;Z)$ gives an accurate estimation of $I(X;Z)$ (\mbox{see Appendix \ref{appendixA}}). Compared to the matrix-based estimator, the binning method gives less accurate results that are violently affected by the resolution of discretization and sample size.
Using this estimator, Yu and Principe visualized the IP trajectories of AEs and suggested the optimal design of AEs based on IP patterns~\cite{IP&AE1}.

The kernel estimator contains a hyperparameter that defines a kernel function of distances between samples. As the estimator depends on the dimension and scale of variables for samples, the hyperparameter should be carefully determined~\cite{IP&AE2}.
Despite careful determination, the matrix-based kernel estimator seems unstable because it is sensitive to the training setup of neural networks. Moreover, once the information process of deep neural networks is quantified by this estimator, it sometimes violates the DPI, which~is a necessary condition for interpreting layer stacks as Markov chains. 
We found that the raw activities of neural networks can result in inaccurate entropy estimations irrespective of the estimation schemes when they have different dimensions and scales depending on layers. Large activities tend to overestimate their entropies, whereas low activities tend to underestimate their entropies. 
In particular, the use of a linear activation function or rectified linear unit (ReLU) often results in the violation of DPI (see Figures 6 and 9 in~\cite{IP&AE2}). 
To address this issue, we unified to use a bounded activation function, i.e., sigmoid function ($f(s) = 1/(1+\exp(-s)))$, except for VAE and LAE, whose bottleneck layers used Gaussian sampling and a softmax function, respectively; and this setup eliminated the DPI~violation.

{Saxe et al. argued that using double-sided saturating activation functions such as $f(s)=\tanh(s)$ trivially induces the simplifying phase on the IP, and it is not related to the generalization of machine learning~\cite{generalIP}. 
They showed that the mutual information, estimated} by the binning method, first increases and then decreases as the weight parameters of neural networks get larger. The second decreasing phase of mutual information causes the simplifying phase. We performed the same task with various activation functions, including sigmoid and ReLU, but we estimated the mutual information using the aforementioned matrix-based kernel method.
Then, we confirmed that the second decreasing phase did not occur by merely increasing the weight parameters, suggesting that the existence of the simplifying phase does not depend on the selection of activation functions in our matrix-based kernel method. Further details on this experiment are provided in \mbox{Appendix \ref{appendixA}}.
For~those who are interested in using IP analysis, we have provided the complete source code and documentation on GitHub~\cite{Github}. 

\section{Results}
\label{results}
{In this section, we examine the MI of various AE models using the method introduced in the previous section, and visualize it on IP. Our main concern is whether the phase transition in IP can be observed in representation learning. Furthermore, by comparing the IPs of different AEs, we investigate how the various techniques we adopted for efficient training of neural networks modified the information flow in latent space.}

\subsection{Information Flows of Autoencoders}
\begin{figure}[t]
\includegraphics[scale=0.55]{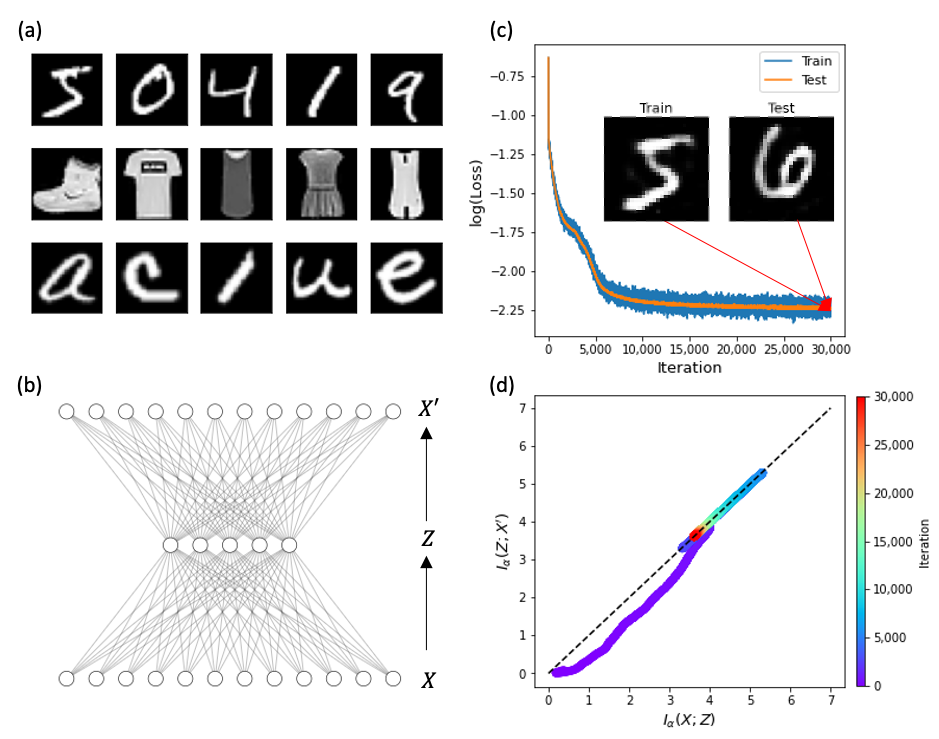}
\caption{Information transmission and compression of autoencoders. (\textbf{a}) Image datasets $X$ of MNIST (top row), Fashion-MNIST (middle), and EMNIST (bottom). (\textbf{b}) Network structure of a shallow autoencoder: input $X$, hidden $Z$, and output $X'$. Note that node numbers are arbitrary for a schematic display. (\textbf{c}) Error (or loss) between desired output $X$ and reconstructed output $X'$ for training (blue) and test (orange) data during learning iterations. Insets are snapshots of reconstructed training and test images of $X'$ at the final iteration. (\textbf{d}) Trajectory of mutual information ($I_\alpha(X;Z), I_\alpha(Z;X')$) on the information plane. The color bar represents the number of iterations.}
\label{figure1}
\end{figure}

In this study, we investigated the information process of representation learning for real image datasets (Figure~\ref{figure1}a): MNIST \cite{MNIST}, Fashion-MNIST \cite{FMNIST}, and EMNIST \cite{EMNIST}. 
MNIST has 60,000 training and 10,000 testing images of $28\times 28$ pixels of 10 hand-written digits (0--9). 
Fashion-MNIST has the same data size as MNIST for 10 different fashion products, such as dresses and shirts. 
Finally, EMNIST is an extension of MNIST; it contains 10 digits and 26 uppercase (A--Z) and lowercase letters (a--z). 
In this paper, we focused on the results of MNIST because the results of Fashion-MNIST and EMNIST are basically the same (refer \cite{Github}). 

\begin{figure*}[t]
\includegraphics[scale=0.7]{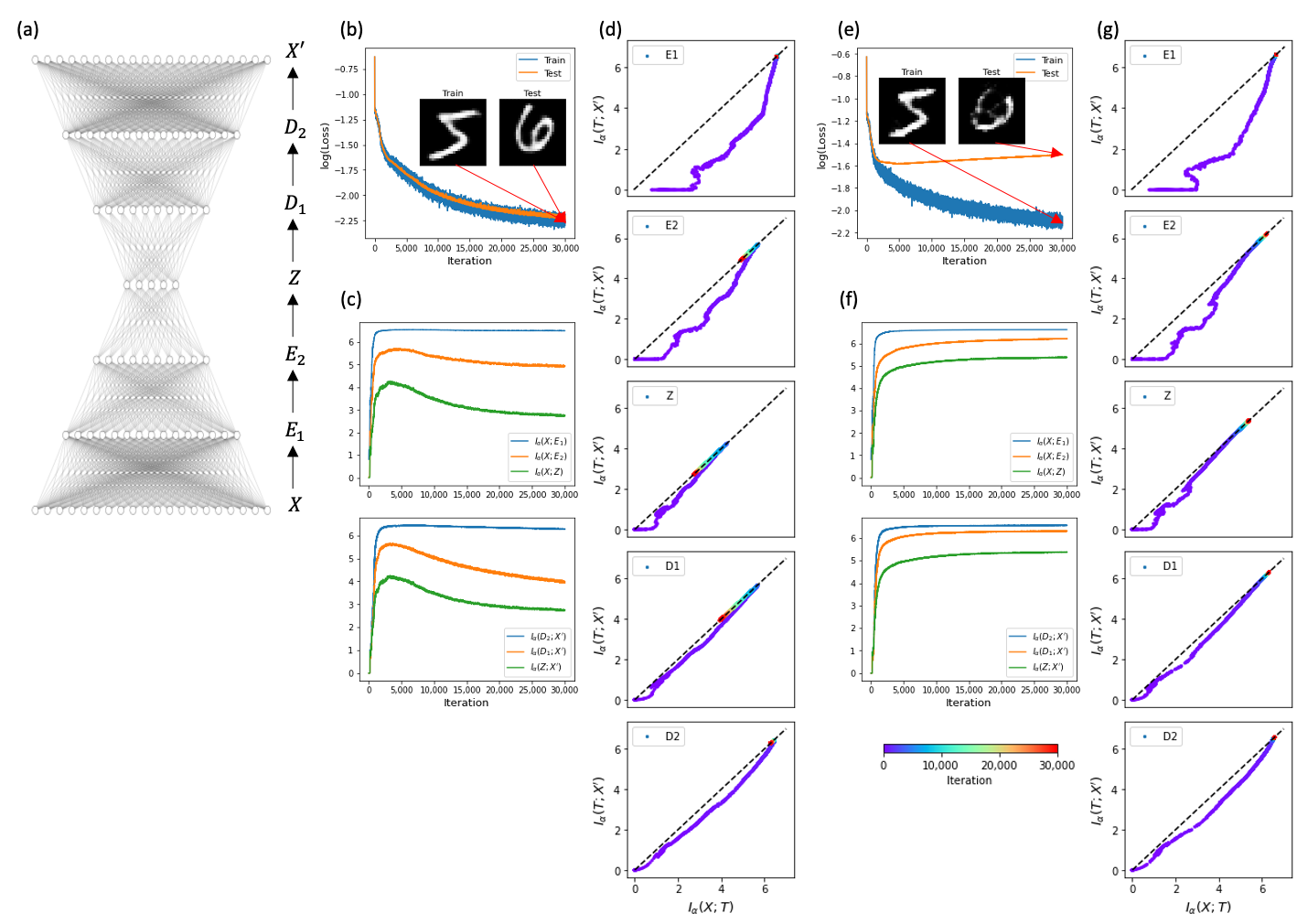}
\centering
\caption{The simplifying phase and generalization of representation learning. (\textbf{a}) A deep autoencoder with input $X$; two~encoders, $E_1$ and $E_2$; bottleneck $Z$; two decoders, $D_1$ and $D_2$; and output $X'$. (\textbf{b}) Learning errors for training (blue) and test (orange) data during iterations. Insets are snapshots of reconstructed training and test images of $X'$ at the final iteration. (\textbf{c}) Changes in input mutual information (upper) and output mutual information (lower) during iterations. (\textbf{d}) Learning trajectories on the information plane. The general variable $T$ stands for $E_1, E_2, Z, D_1$, or $D_2$. (\textbf{b}--\textbf{d}) Experiments with the full training set of 60,000 MNIST data. (\textbf{e}--\textbf{g}) Experiments with the 10\% training set of 6,000 MNIST data.}
\label{figure2}
\end{figure*}

For representation learning of MNIST, we first considered a shallow AE ($X-Z-X'$) that included a single hidden or bottleneck layer (Figure~\ref{figure1}b). The input, hidden, and~output layers had $n_X = 28 \times 28 = 784, n_Z = 50$, and $n_{X'} = 784$ nodes, respectively. 
We~considered a fully-connected network between layers with the loss functions listed in Table~\ref{AE type}.
For~the optimization of network weights, we used the stochastic gradient descent method with Adam optimization, given a batch size of 100 for a total of 50 epochs.
With each learning iteration, the MSE$(X, X')$ kept decreasing (Figure~\ref{figure1}c). This implies that the output $X'$ of AE successfully reproduced the input image $X$ of training data.
To measure the generalization ability of the AE, we examined the reproduction ability of the AE for test images that were not used in the learning process.
We confirmed that the test error was as small as the training error. Given the faithful reproduction of input images, the indifferent error between training and test images defines successful generalization as usual.
The~IP trajectory of the AE during the learning process is presented in Figure~\ref{figure1}d.
As expected, the~trajectory satisfies the inequality of $I_\alpha(X;Z) \geq I_\alpha(Z;X')$, and ended up on their equality line because of $X' \approx X$ at the end of training.
As observed by Shwartz and Ziv~\cite{IP}, the~IP trajectory showed two distinct phases of fitting and simplifying.
In the initial fitting phase, the~input mutual information $I_\alpha(X;Z)$ between $X$ and $Z$ increased. Then, during the second simplifying phase, $I_\alpha(X;Z)$ decreased.
Note that this representation learning showed a simultaneous decrease in the output mutual information, whereas general supervised learning maintained the output mutual information as constant during the simplifying~phase. 

Next, we considered a deep AE ($X-E_1-E_2-Z-D_1-D_2-X'$) with two additional encoder layers before the bottleneck layer and two decoder layers after the bottleneck layer (Figure~\ref{figure2}a).
The corresponding node numbers for the inner layers were $n_{E_1} = 256$, $n_{E_2} = 128, n_{Z} = 50, n_{D_1} = 128$, and $n_{D_2} = 256$.
The deep AE exhibited similar learning accuracy and generalization ability to the shallow AE (Figure~\ref{figure2}b). 
During the learning process, we~measured the mutual information using the matrix-based kernel estimator and confirmed that the learning process of the deep AE satisfied the DPI (Figure~\ref{figure2}c).
We~observed the simplifying phase in the inner layers of $E_2, Z$, and $D_1$ (Figure~\ref{figure2}d). However, the simplifying phase was not evident in the outer layers of $E_1$ and $D_2$ that had relatively large dimensions with high information capacity. 



Subsequently, we explored whether the simplifying phase appeared even with a small amount of training data.
Unless sufficient training data are provided, machine learning can easily overfit a small amount of training data and fail to generalize the test data.
We~conducted a learning experiment with the deep AE using 10\% of the total training data.
The training error kept decreasing, similarly to the training error given the full training data. However, the test error was significantly larger than the training error (Figure~\ref{figure2}e). 
This demonstrates that the deep AE failed to generalize.
After confirming the satisfaction of DPI (Figure~\ref{figure2}f), we examined the IP trajectories.
Unlike the results of full training data, we did not observe the simplifying phase from any layers (Figure~\ref{figure2}g).
Thus, this difference suggests that the simplifying phase seems to be associated with generalization by removing irrelevant details.

\subsection{Sparse Activity and Constrained Weights}
\begin{figure}[t]
\includegraphics[scale=0.5]{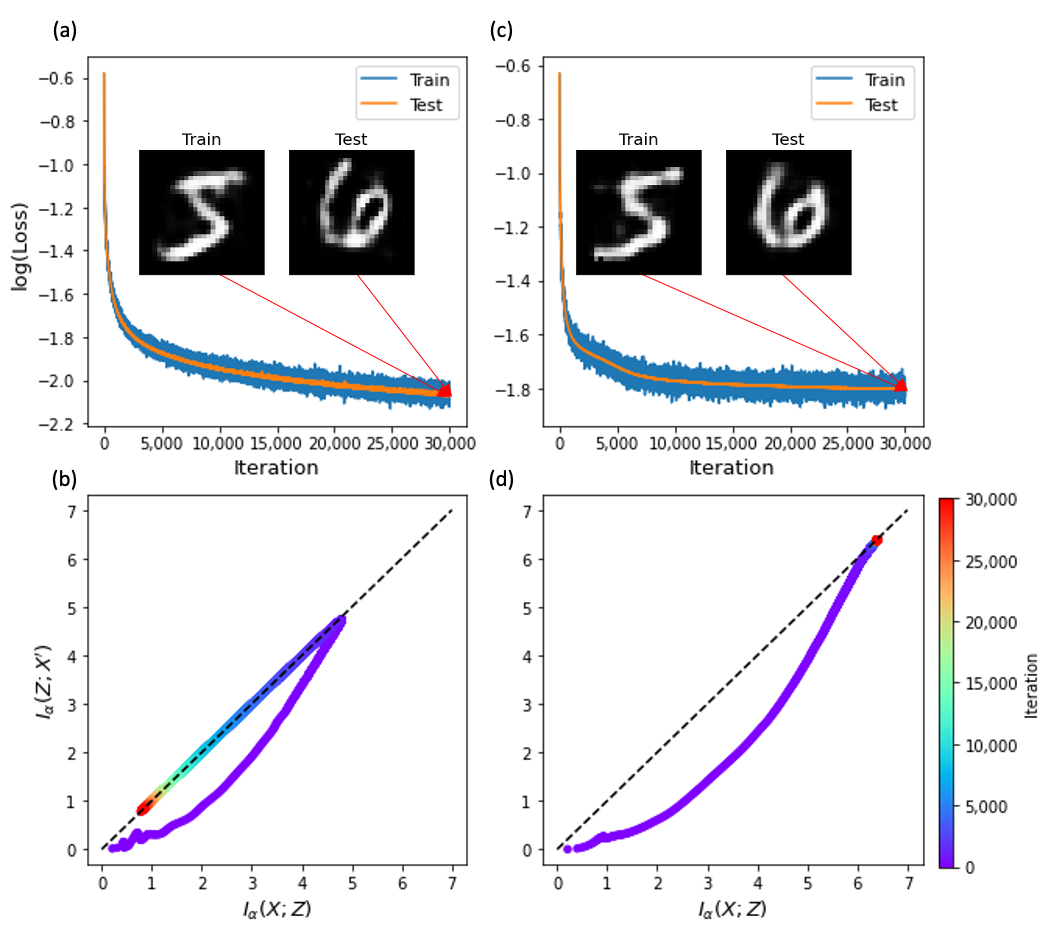}
\caption{Information compression in constrained autoencoders. (\textbf{a}) Learning errors for training (blue) and test (orange) data during iterations. Insets are snapshots of reconstructed training and test images of $X'$ at the final iteration. (\textbf{b}) Learning trajectories on the information plane. (\textbf{a},\textbf{b}) Results of sparse autoencoders (SAE). (\textbf{c},\textbf{d}) Results of tied autoencoders (TAE). The network structure of SAE and TAE can be represented by $X-Z-X'$.}
\label{figure3}
\end{figure}

To examine the effect of regularization on the information flows, we considered different species of AEs that can modify the learning phases.
SAE and TAE have additional regularization phases for node activities and weight parameters, respectively, in comparison to vanilla AE (Table \ref{AE type}).
First, we examined SAE, which has the same structure as a shallow AE. 
SAE showed perfect learning and generalization (Figure~\ref{figure3}a). 
It is of particular interest that the simplifying phase is markedly exaggerated in SAE (Figure~\ref{figure3}b).
The sparsity penalty of SAE turns off unnecessary activities of hidden nodes, which can accelerate the simplifying phase.

Second, we examined TAE, which also has the same structure as shallow AE and SAE, but has a weight constraint of $W_E = W_D^T$.
Similarly to shallow AE and SAE, TAE showed perfect learning and generalization; 
however, its learning accuracy was slightly lower under the weight constraint (Figure~\ref{figure3}c).
However, TAE did not exhibit the simplifying phase (Figure~\ref{figure3}d).
This implies that the simplifying phase is not necessary for generalization.
Given the weight constraint, TAE seems to have less potential capacity to remove irrelevant information than vanilla AE.

\subsection{Constrained Latent Space}
Now, we survey another species of AEs that more actively modify the latent space of the bottleneck layer, and further investigate the information flows in the learning process. 

\begin{figure*}[t]
\includegraphics[scale=0.8]{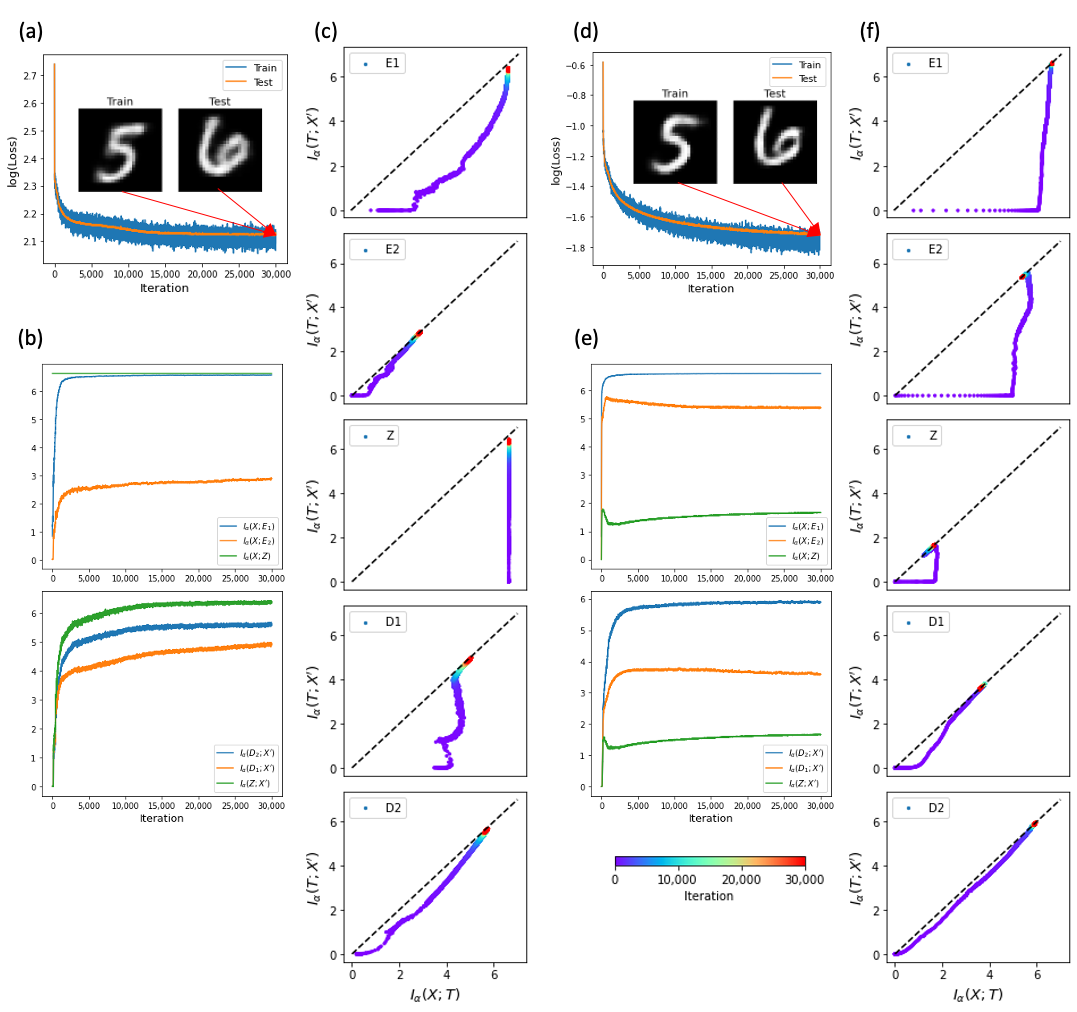}
\caption{Information trajectories of generative models. (\textbf{a}) Learning errors for training (blue) and test (orange) data during iterations. Insets are snapshots of reconstructed training and test images of $X'$ at the final iteration. (\textbf{b}) Changes in the input mutual information (upper) and output mutual information (lower) during iterations. (\textbf{c}) Learning trajectories on the information plane. (\textbf{a}--\textbf{c}) Results of a variational autoencoder (VAE). (\textbf{d}--\textbf{f}) Results of a label autoencoder (LAE). VAE and LAE had a deep network structure with $X-E_1-E_2-Z-D_1-D_2-X'$. The general variable $T$ denotes $E_1, E_2, Z, D_1$, or $D_2$.}
\label{figure4}
\end{figure*}

VAE is a generative model that maps input data $X$ into a Gaussian distribution $q_\phi(Z|X)$ for the latent variable $Z$.
We considered a deep VAE that had the same structure ($X-E_1-E_2-Z-D_1-D_2-X'$) as deep AE, and confirmed that the VAE can learn the training data of MNIST and generalize to reproduce the test data (Figure~\ref{figure4}a).
However, because VAE had a special constraint for the bottleneck layer, the information process from the input layer into the bottleneck layer did not satisfy the DPI (Figure~\ref{figure4}b).
The~mutual information $I_\alpha(X;Z)$ between $X$ and $Z$ did not change during the training process.
Indeed, the fixed value was close to the maximum entropy of $X$ given its batch size of 100 samples, $I_\alpha(X;Z) \approx \log_2(100) \approx 6.6$, which was independent of dimension $n_Z$ of $Z$ (data not shown).
It is noteworthy that the mutual information between $X$ and $Z$ did not change for the learning process, although the mapping $X \rightarrow Z$ kept reorganizing to distinguish the feature differences of $X$. This shows a limitation of the information measure $I_\alpha(X;Z)$, which failed to capture the content-dependent representation of $Z$.
Besides the bottleneck layer, other layers still satisfied the DPI.
Next, we displayed the IP trajectories of VAE for each layer (Figure~\ref{figure4}c). We did not observe the simplifying phase in any layer in the VAE. Therefore, VAE can generalize without the simplifying phase, similarly to TAE.


LAE is another generative model that maps $X$ into $Z$, where $Z$ corresponds to label $Y$ of $X$.
Thus, unlike other AE models, LAE uses label information to shape its latent space.
We used the same deep network structure as the deep AE and VAE for LAE.
The~deep LAE could also learn the training data of MNIST and generalize to reproduce the test data as well (Figure~\ref{figure4}d). 
LAE satisfied the DPI (Figure~\ref{figure4}e), and its IP trajectories also satisfied the ineqaulity of $I_\alpha(X;T) \geq I_\alpha(T;X')$ (Figure~\ref{figure4}f).
It is interesting that LAE has orthogonal learning phases.
LAE first increased the input mutual information $I_\alpha(X;T)$ for the encoding part. Once LAE arrived at a certain maximal $I_\alpha(X;T)$, the output mutual information $I_\alpha(T;X')$ started to increase.
This shows that LAE first extracts information from the input data relevant for the label classification, and then transfers information to output for reproducing input images.
We found that the LAE did not exhibit the simplifying phase, even though it successfully generalized.

\section{Discussion}
\label{discussions}
We studied the information flows in the internal representations of AEs using a matrix-based kernel estimator.
AEs are perfect models to investigate how the information flows are shaped during the learning process depending on the network designs, since they have diverse species with various depths, sparsities, weight constraints, and hidden representations.
When we used sufficient training data, shallow and deep AEs demonstrated the simplifying phase, following the fitting phase, along with the generalization ability to reproduce test data, thereby confirming the original proposal by Shwartz-Ziv and Tishby~\cite{IP}. However, when we used a small amount of training data to induce overfitting, the AEs did not exhibit a simplifying phase and generalization, suggesting that the simplifying phase is associated with generalization.
When a sparsity constraint was imposed in the hidden activities of SAE, regularization amplified the simplifying phase and provided more efficient representations for generalization.
However, the constraining weight parameters ($W_E = W_D^T$) of TAE showed perfect generalization in the absence of the simplifying phase. Furthermore, VAE and LAE, shaping the latent space with a variational distribution and label information, also achieved generalization without the simplifying phase.
These~counterexamples of TAE, VAE, and LAE clearly demonstrate that the simplifying phase is not necessary for the generalization of models. 

It is noteworthy that the absence of the simplifying phase does not mean that compression does not occur in representation learning. When the encoder part has a narrowing architecture, information compression is inevitable, as demonstrated by the DPI. Then, the~removal of irrelevant information from data may contribute to the generalization of models.
After the completion of representation learning, AEs obtain a certain amount of mutual information $I_\alpha(X;Z) = I_{final}$ between the input data $X$ and its internal representation $Z$.
The paths that obtain $I_{final}$ seem different between AEs.
In TAE, VAE, and~LAE, $I_\alpha(X;Z)$ monotonically increases to $I_{final}$.
However, in vanilla AE and SAE, $I_\alpha(X;Z)$ first increases beyond $I_{final}$, and then decreases back to $I_{final}$. The backward process is called the simplifying phase.  
As the loss function of representation learning never includes any instruction for the path of $I_\alpha(X;Z)$, it is not surprising that the existence of the simplifying phase is not universal.
{In summary, in the basic structure of AE, we found that the simplifying phase is related to the generalization of the model, and confirmed that learning dynamics of neural network can be interpreted with IB theory. However, for~several variants of AE, no simplifying phase was observed, suggesting that all types of deep learning do not follow a universal learning dynamics.}

{Although observations and physical meanings of the phase transition in IP were contradictory in previous studies, it is still manifest that IP analysis is an excellent tool to monitor information transmission and compression inside the ``black box'' of neural networks.}
For IP analysis, accurate information estimation is a prerequisite. In general, it is difficult to calculate the entropies of high-dimensional variables, but we could solve this problem by estimating the physical quantities corresponding to the entropies in a kernel space. When we applied the estimator to the representation learning, we found that it is critical to use bounded activation functions. 
When we used ReLU as an activation function, the DPI was easily violated, although we observed the simplifying phase in this setting. Thus, it can be problematic to estimate the mutual information from unbounded variables with different scales for different layers.
In this study, we provided concrete grounds to further explore the theoretical understanding of information processing in deep learning.

\section*{Acknowledgment}
This work was supported in part by the National Research Foundation of Korea (NRF) grant (grant number 2018R1A2B6004914) (S.L.), the New Faculty Startup Fund from Seoul National University, and the NRF grant funded by the Korea government (MSIT) (grant number 2019R1F1A1052916) (J.J.).

\setcounter{figure}{0}
\renewcommand{\thefigure}{A\arabic{figure}}
\appendix

\begin{figure*}[t]
\includegraphics[scale=0.5]{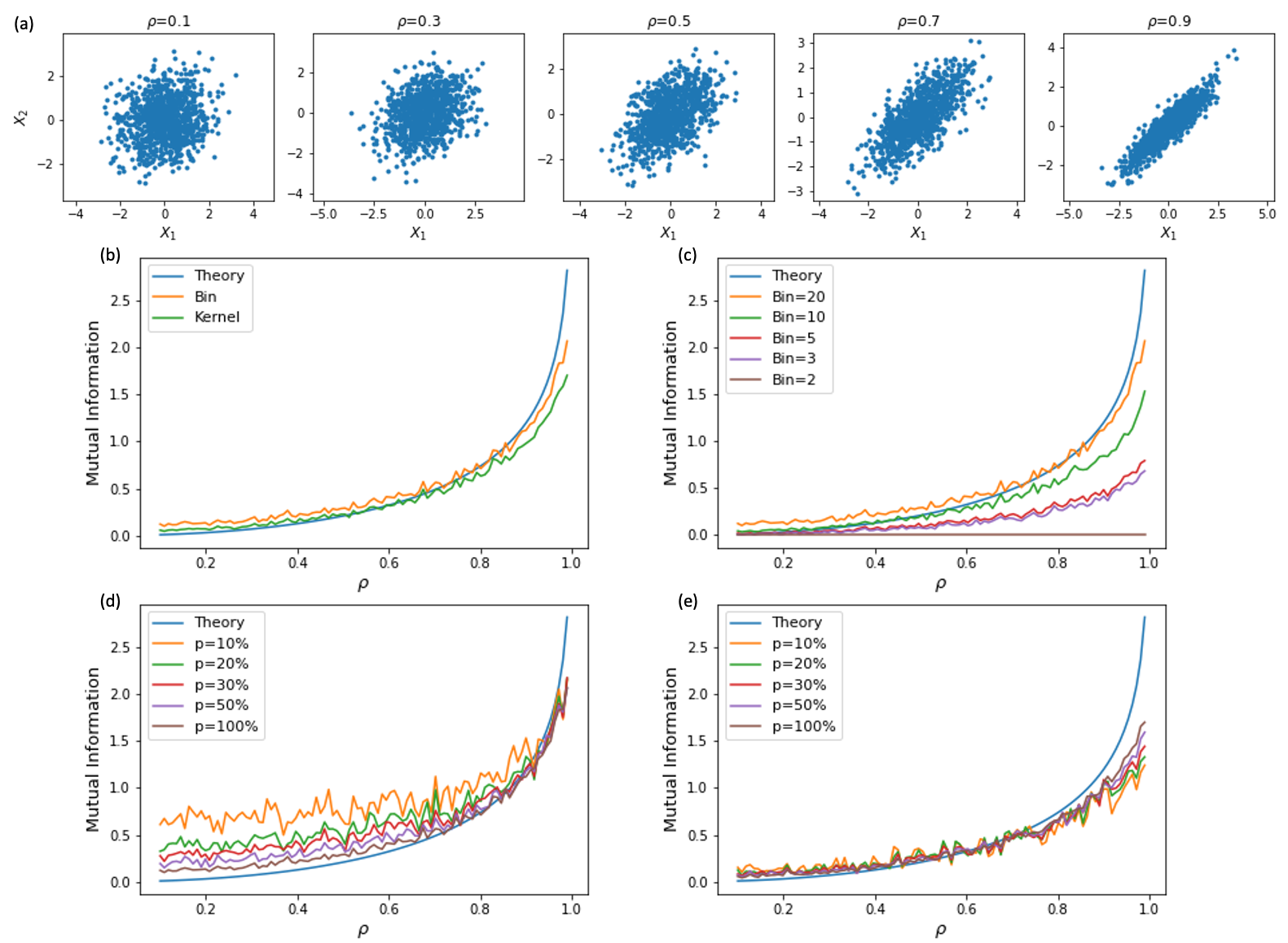}
\caption{Estimation of mutual information by binning and matrix-based kernel methods. (\textbf{a}) Distributions of two variables $(X_1, X_2)$ following bivariate normal distributions with various correlation strengths of $\rho$. (\textbf{b}) Exact mutual information (Theory) and its optimal estimation by the binning method (Bin) and the matrix-based kernel method (Kernel). (\textbf{c}) Mutual information estimated by the binning method with various binning levels (Bin) of discretization for the continuous variables $X_1$ and $X_2$. Mutual information estimation with various sample sizes (p, percentage of the sample size to the entire data) of (\textbf{d}) the binning method and (\textbf{e}) the matrix-based kernel method.} 
\label{figure5}
\end{figure*}

\begin{figure}[h]
\includegraphics[scale=0.38]{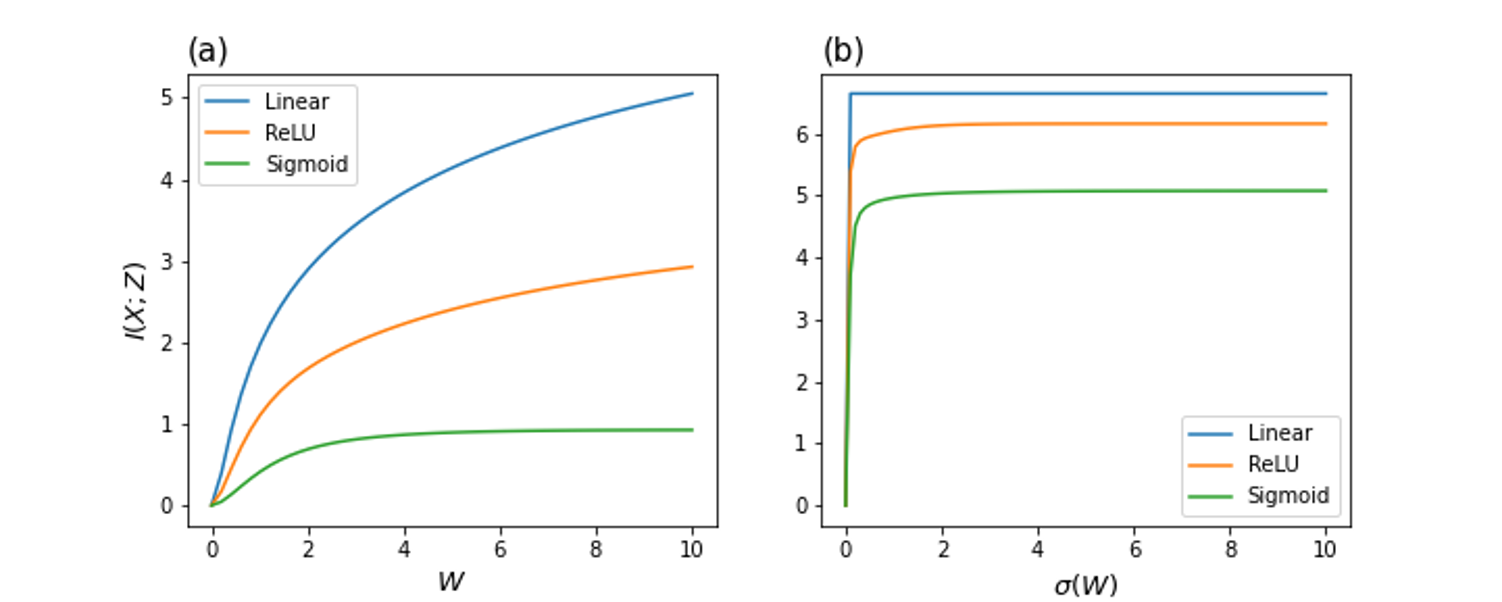}
\caption{Mutual information obtained by matrix-based kernel estimation. (\textbf{a}) Input mutual information in a three-neuron network ($x-z-y$). Input $x$ was sampled from a standard normal distribution and the hidden activity was computed by $z = f(wx)$, where $w$ is the weight and $f(s)$ is an activation function. Three different activation functions (linear, ReLU, and sigmoid) were used. (\textbf{b}) Input mutual information for a general setup with 100-dimensional input vector $X$ and 50-dimensional hidden vector $Z=f(WX)$. In this setup, weight $W$ was a $50\times100$ matrix whose elements were sampled from a uniform distribution.} 
\label{figure6}
\end{figure}

\section{Matrix-Based Kernel Estimator of Mutual Information}\label{appendixA}
The matrix-based kernel method \cite{kernel} estimates R\'enyi's $\alpha$-entropy for a random variable $X$ by
\begin{eqnarray}
    H_{\alpha}(X)=\frac{1}{1-\alpha}\int_{x\in\mathcal{X}} f^{\alpha}_X(x)dx. \label{Renyi entropy definition} 
\end{eqnarray}

Let $X=\left\{x_1, x_2, ..., x_N \right\}$ denote $N$ data points and $\kappa:\mathcal{X} \times \mathcal{X} \rightarrow \mathbb{R}$ be a real valued positive definite kernel that defines a Gram matrix $K \in \mathbb{R}^{N\times N}$ as $K_{ij}=\kappa(x_i,x_j)$. The~normalized Gram matrix is defined as
\begin{eqnarray}
    A_{ij}=\frac{1}{N}\frac{K_{ij}}{\sqrt{K_{ii}K_{jj}}}. \label{normalized Gram matrix}
\end{eqnarray}

Then, the matrix-based R\'enyi's $\alpha$-order entropy is given by
\begin{eqnarray}
    S_\alpha(A) &=& \frac{1}{1-\alpha}\log_2\left[\text{tr}(A^\alpha)\right] \nonumber \\
    &=&\frac{1}{1-\alpha}\log_2\left[\sum_{i=1}^N\lambda_i(A)^\alpha\right], \label{Renyi entropy_appendix}
\end{eqnarray}
where $\lambda_i(A)$ denotes the $i$-th eigenvalue of $A$. In the limit of $\alpha\rightarrow 1$, Equation~(\ref{Renyi entropy_appendix}) is reduced to the Shannon entropy-like object
\begin{eqnarray}
    \lim_{\alpha\rightarrow 1}S_\alpha(A) = -\sum_{i=1}^N\lambda_i(A)\log_2 \lambda_i(A).
\end{eqnarray}

We used $\alpha=1.01$ in this study.
The joint entropy of two random variables $X$ and $Z$ can be defined as
\begin{eqnarray}
    S_\alpha(A,B)=S_\alpha\left(\frac{A\circ B}{\text{tr}(A\circ B)}\right), \label{joint entropy_appendix}
\end{eqnarray}
where $A$ and $B$ are Gram matrices of $X$ and $Z$, respectively, and $A\circ B$ denotes the Hadamard product. From Equations~(\ref{Renyi entropy_appendix}) and (\ref{joint entropy_appendix}), the mutual information in the kernel space is defined as
\begin{eqnarray}
    I_\alpha(X;Z)=S_\alpha(A)+S_\alpha(B)-S_\alpha(A,B). \label{mutual information_appendix}
\end{eqnarray}

The Gaussian kernel is commonly used:
\begin{eqnarray}
    \kappa_\sigma(x_i,x_j)=\exp\left(-\frac{||x_i-x_j||_F^2}{2\sigma^2}\right), \label{Gaussian kernel}
\end{eqnarray}
where $||\cdot||_F$ denotes the Frobenius norm. 
There are crucial factors that affect the estimation performance, such as the Gaussian kernel bandwidth $\sigma$ and the scale and dimension of kernel input. The asymptotic behavior of entropy by varying $\sigma$ can be denoted by
\begin{align}
\lim_{\sigma\rightarrow 0}S_\alpha(A)&=\log N\\
\lim_{\sigma\rightarrow \infty}S_\alpha(A)&=0. \label{asymptotic behavior}
\end{align}

Large-scale and high-dimensional features of input have the same effect as a small $\sigma$---the overestimation of entropy.
In contrast, small-scale and low-dimensional features of input give the same effect as large $\sigma$, which results in the underestimation of entropy. Therefore, proper hyperparameter tuning is required for $\sigma$ to avoid excessively high or low saturation of entropy during training. Scott's rule \cite{Scott}, a simplified version of Silverman's rule \cite{Silverman}, is commonly used for selecting the width of Gaussian kernels:
\begin{eqnarray}
\label{Scott}
    \sigma=\gamma N^{-1/(4+n)},
\end{eqnarray}
where $\gamma$ is an empirically determined constant. As Equation~(\ref{Scott}) is a monotonically increasing function with respect to feature dimension $n$, it compensates for higher feature dimension. We used $\gamma=2$ for our experiments. 

To validate the matrix-based kernel method, we consider a bivariate normal distribution as a simple example. Let us assume that two variables $X_1$ and $X_2$ follow a bivariate normal distribution:
\begin{align}
    \begin{pmatrix}
    X_1 \\
    X_2
    \end{pmatrix}
    \sim \mathcal{N}
    \begin{pmatrix}
    \begin{pmatrix}
    \mu_1 \\
    \mu_2
    \end{pmatrix}
    , \Sigma
    \end{pmatrix}
    ,\quad \Sigma=
    \begin{pmatrix}
    \sigma_1^2 & \rho\sigma_1\sigma_2 \\
    \rho\sigma_1\sigma_2 & \sigma_2^2
    \end{pmatrix},
\end{align}
where $\mu_i$ and $\sigma_i$ are mean and standard deviation of the variable $X_i\:(i=1,2)$, respectively, and $\rho$ denotes their correlation strength. The entropy of each variable and their joint entropy are given as follows: 
\vspace{6pt}
\begin{align}
    H(X_i)&=\frac{1}{2}\log(2\pi e\sigma_i^2),\\
    H(X_1,X_2)&=\frac{1}{2}\log((2\pi e)^2|\Sigma|) \nonumber \\
    &=\log(2\pi e\sigma_1\sigma_2)+\frac{1}{2}\log(1-\rho^2).
\end{align}

Then, the mutual information between $X_1$ and $X_2$ can be exactly computed as 
\begin{align}
    I(X_1;X_2) &=H(X_1)+H(X_2)-H(X_1,X_2) \nonumber \\
    &=-\frac{1}{2}\log(1-\rho^2).
\end{align}

Now, we estimated $I(X_1;X_2)$ numerically using a binning method and the matrix-based kernel method with 1,000 samples generated from the bivariate normal distribution of mean 0 and variance 1.
Figure~\ref{figure5}a shows $(X_1, X_2)$ distributions under different correlation strengths. As shown in figure~\ref{figure5}b, the theoretical value of $I(X_1;X_2)$ is consistent with the estimated values of the binning method with a proper quantizer (Bin = 20) and the matrix-based kernal method with a proper hyperparameter ($\gamma=2$).
Note~that Bin represents the level of discretization for the continuous activity of $X_i$.
For the binning method, \mbox{Figure~\ref{figure5}c,d} show that its estimate of mutual information largely varies depending on the binning level and sample size. However, the matrix-based kernel method gives a robust estimate relatively less sensitive to the sample size.

Saxe {\it et al.} observed that information estimation depends on the activation functions in a simple setup of a three neuron model ($x-z-y$)~\cite{generalIP}.
They sampled a scalar input $x$ from a standard normal distribution of $\mathcal{N}(0,1)$ and multiplied it by a constant weight $w$; subsequently, they determined the hidden activity $z = f(wx)$ using a nonlinear activation function $f(s)$.
Then, they discretized $z$ and estimated the input mutual information $I(x;z)$ using a binning method. 
When the unbounded activation function of $f(s)=\textrm{ReLU}(s)$ was used, $I(x;z)$ continued to increase with $w$.
However, when the bounded activation function of $f(s)=\tanh(s)$ was used, $I(x;z)$ first increased with $w$, and then decreased as $w$ increased.
This is a natural result when the binning method is used to estimate the mutual information because large activities are saturated with large $w$ (Figure 2 in~\cite{generalIP}).
We analyzed the same task with the matrix-based kernel method (Figure~\ref{figure6}a). 
When~the activation function is a sigmoid function of $f(s)=1/(1+\exp(-s))$, $I(x;z)$ does not decrease at large $w$; however, the absolute value looks different from the unbounded activation functions of linear ($f(s)=s$) and ReLU ($f(s)=\textrm{ReLU}(s)$).
We also considered a more complex network with 100-dimensional input $X$ sampled from $\mathcal{N}(0,1)$.
In this case, weight $W$ was represented by a $50\times100$ matrix whose elements were sampled from a uniform distribution of $\mathcal{U}(0,1)$; then, the hidden activity $Z$ becomes a 50-dimensional vector, i.e., $Z=f(WX)$. We then observed $I_\alpha(X;Z)$ while increasing the standard deviation of weight $W$ (Figure~\ref{figure6}b).
We confirmed that $I_\alpha(X;Z)$ at large $W$ does not decrease when a sigmoid activation function is used, like the unbounded activation functions of linear and ReLU. 
Therefore, this experiment demonstrated that the matrix-based kernel method is a robust estimation technique for bounded activation function and the simplifying phase cannot be attributed to the selected activation function.

\begin{figure*}[t]
\includegraphics[scale=0.7]{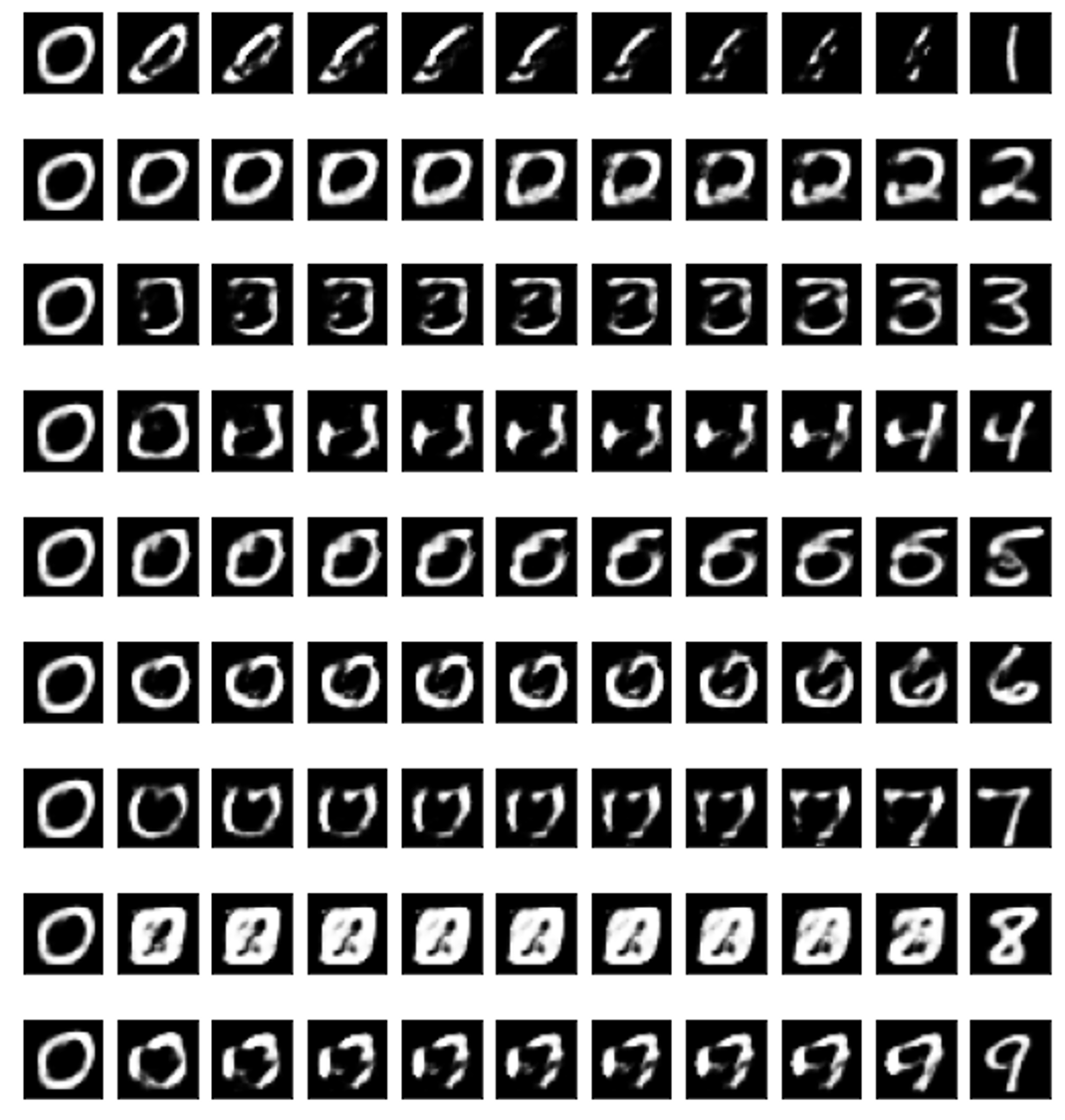}
\caption{Manifold learning in LAE. This represents the interpolation image when one-hot vector corresponding to zero; i.e., $Z_0$ = $[1,0,0,\cdots,0]$, is transformed to other digits as an input of the LAE decoder part. For instance, the first row represents the reproduction $X'$ decoded from $Z = a Z_0 + (1-a)Z_1$ by decreasing $a$ from 1 to 0.}
\label{figure7}
\end{figure*}

\section{LAE: Label Autoencoder}\label{appendixB}
LAE is a generative model that shapes its latent space using label classification. The~explicit form of the LAE loss function is given as follows:
\begin{eqnarray}
    \mathcal{L}_{\textrm{LAE}} = \frac{1}{N}\sum_{i=1}^N||X_i-X_i'||^2 - \frac{\lambda}{N} \sum_{i=1}^N\sum_{j=1}^{n_Z} Y_{i,j}\log Z_{i,j},
\end{eqnarray}
where $N$ is the batch size and $n_Z$ is the feature dimensionality of the label. 
$Z_{i,j}$ is the softmax output of encoder that predicts the $j$-th class of the $i$-th sample and $Y_{i,j}$ is the corresponding true label. The first term is a reconstruction error (MSE) and the second term is a regularization given by the classification error of the encoder. The regularization coefficient $\lambda$ is set to $0.01$. Figure~\ref{figure7} shows the manifold learning of LAE when the one-hot vector corresponding to zero is changed to other digits as an input of the decoder. It is a two-dimensional submanifold embeded in a ten-dimensional label latent space.

\bibliographystyle{apsrev4-1}
\bibliography{references}

\end{document}